\documentclass[conference]{IEEEtran}
\usepackage[left=0.75in,right=0.75in,top=1.0in,bottom=0.75in]{geometry}
\IEEEoverridecommandlockouts
\usepackage[utf8]{inputenc} %
\usepackage[T1]{fontenc}    %
\usepackage{hyperref}       %
\usepackage{url}            %
\usepackage{booktabs}       %
\usepackage{amsfonts}       %
\usepackage{nicefrac}       %
\usepackage{microtype}      %
\usepackage{xcolor}         %
\usepackage{siunitx}

\usepackage{cite}
\usepackage{amsmath,amssymb,amsfonts}
\usepackage{algorithmic}
\usepackage{graphicx}
\usepackage{textcomp}
\usepackage[font=footnotesize]{subcaption}
\usepackage{xcolor}
\def\BibTeX{{\rm B\kern-.05em{\sc i\kern-.025em b}\kern-.08em
    T\kern-.1667em\lower.7ex\hbox{E}\kern-.125emX}}
\begin{document}

\title{Goal-based Trajectory Prediction for improved Cross-Dataset Generalization\\
}
\author{
Daniel~Grimm$^{1}$,
Ahmed Abouelazm$^{1}$,
and J.~Marius~Zöllner$^{1}$
\thanks{$^{1}$FZI Research Center for Information Technology, Karlsruhe, Germany
{\tt\small \{grimm, abouelazm, zoellner\}@fzi.de}}%

}%

\maketitle

\begin{abstract}
To achieve full autonomous driving, a good understanding of the surrounding environment is necessary. Especially predicting the future states of other traffic participants imposes a non-trivial challenge. Current SotA-models already show promising results when trained on real datasets (e.g. Argoverse2\cite{argoverse2}, NuScenes\cite{nuscenes}). Problems arise when these models are deployed to new/unseen areas. Typically, performance drops significantly, indicating that the models lack generalization. In this work, we introduce a new Graph Neural Network (GNN) that utilizes a heterogeneous graph consisting of traffic participants and vectorized road network. Latter, is used to classify goals, i.e. endpoints of the predicted trajectories, in a multi-staged approach, leading to a better generalization to unseen scenarios. We show the effectiveness of the goal selection process via cross-dataset evaluation, i.e. training on Argoverse2\cite{argoverse2} and evaluating on NuScenes\cite{nuscenes}.

\end{abstract}


\section{Introduction}
Automated shuttle buses and automated freight transportation could have great benefits for our society, provided they have a safe highly automated driving (HAD) function.
Thereby it is important to reliably predict the future behavior of surrounding traffic participants.
Such predictions are crucial for anticipating potential interactions and ensuring safe maneuver planning in complex environments.
The current state-of-the-art (SotA) utilizes machine-learning techniques like Transformers or GNNs.
Models like Wayformer \cite{wayformer} and QC-Net \cite{qcnet} exemplify this trend, achieving strong performance on publicly available benchmarks such as the Argoverse2 dataset \cite{argoverse2}.

However, as Unitraj\cite{unitraj} points out, current models perform significantly worse when transferred to datasets that were not used during training, e.g., training on Argoverse2\cite{argoverse2} and validating on NuScenes\cite{nuscenes}.
For example, MTR \cite{mtr} drops $80\%$, i.e. from $2.08~\text{m}$ to $3.72~\text{m}$, regarding the brier-minFDE metric.
Performing poorly on unseen data indicates that these models are not as generalized as the performance on a single dataset might suggest.
Accordingly, the safety of deploying such prediction models within HAD functions remains questionable.

\begin{figure}[t]
  \centering
  \includegraphics[width=0.95\columnwidth]{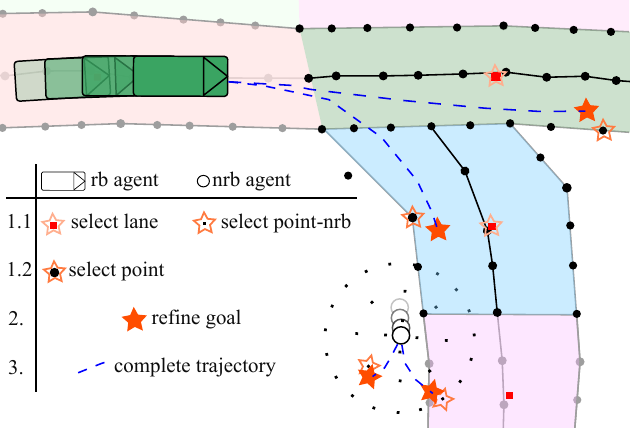}
  \caption{Schematic illustration of the multi-stage goal selection. It shows two predicted trajectories for a vehicle and a pedestrian after firstly selecting potential goals on the map. A step-by-step instruction is given in the lower left corner.}
  \label{fig:overview}
\end{figure}

We argue that generalization across datasets is an important topic and therefore introduce a new model, "HoliGraph:Goal," which is an encoder-decoder GNN with a goal selection process that softly binds predicted trajectories to the road network, leading to lower off-road rates and therefore more human-like trajectories.
It utilizes a heterogeneous graph consisting of agents, lanes, and map-points.
Similar to \cite{hdgt, gorela, qcnet}, features of the graph are encoded in a relative fashion, resulting in translational and rotational invariance of the graph.
To achieve multi-modality, multiple agent-query-nodes per agent are leveraged in the decoder.
In the multi-stage goal selection, a distinction is made between road-bound (rb) agents, e.g., vehicles, cyclists, trucks, etc., and non-road-bound (nrb) agents, e.g., pedestrians.
Rb agents first select a lane and afterwards a point on the lane. Nrb agents directly select an artificially created point-nrb, since they typically attend less to the road.
Afterwards the goal is locally refined and a trajectory to that goal is completed, see Fig.~\ref{fig:overview}.
Our model, HoliGraph: Goal, is able to predict multiple trajectories for all agents in the scene at the same time, giving it real-time capabilities.
The contributions of this paper include:
\begin{itemize}
    \item \textbf{Multi-stage goal selection}: Binding trajectories softly to the road network for increased generalization in cross-dataset evaluation and better performance on smaller datasets like NuScenes \cite{nuscenes}. %
    \item \textbf{Modular heterogeneous graph}: Extending previous work \cite{holigraph} by adding lane nodes and agent-query nodes to a rotational and translational invariant graph. %
\end{itemize}
\begin{figure*}[t]
    \vspace{4pt}
    \centering
    \includegraphics[width=\textwidth]{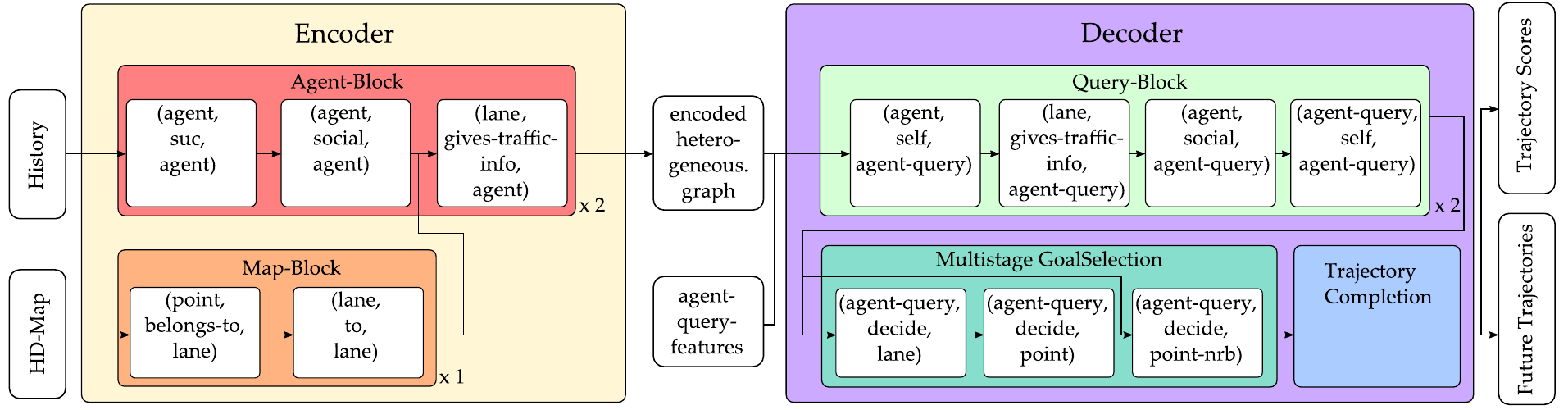}
    \caption{Overview of HoliGraph:Goal. Inputs are embedded in separate embedding modules. The encoder is used to generate a latent representation of all nodes in the scene. The decoder generates meaningful query features, used to select goals on the vectorized map representation. Finally goals are used to predict future Trajectories.}
    \label{fig:concept}
\end{figure*}

\section{Related Work}

Trajectory prediction has seen significant advancements with the transition from dense raster image \cite{covernet} inputs to sparser, vectorized representations, as demonstrated by methods such as LaneGCN \cite{lanegcn} and VectorNet \cite{vectornet}. Vectorized representations not only reduce data redundancy but also enable the explicit modeling of relationships between different entities, leading to more structured motion forecasting.

While vectorized representations have improved the interpretability and scalability of trajectory prediction, significant challenges remain. In particular, many SotA models (e.g., MTR \cite{mtr}, Wayformer \cite{wayformer}) still struggle to generalize across different datasets, as empirically demonstrated in UniTraj \cite{unitraj}, which highlights a strong tendency to overfit to the specific characteristics of the training distribution. This lack of generalization can be further noticed by poor robustness when models are exposed to unseen or adversarial scenarios. For instance, SceneAttack \cite{sceneattack} introduces physically feasible perturbations to road topologies, under which SotA models such as LaneGCN \cite{lanegcn} exhibit implausible or off-road trajectories in up to 60\% of scenarios. These limitations underscore the need for trajectory prediction frameworks that not only leverage structured representations but also prioritize generalization and resilience to distributional shifts.

Recently, researchers have proposed the concept of goal points to guide trajectory predictions and improve generalization and robustness. TNT \cite{tnt} introduced a goal-conditioned prediction framework by employing a heuristic-driven sampling algorithm to generate candidate goal points. The model learns to score these candidates, and non-maximum suppression (NMS) is applied to select a final subset of goal points, for which complete trajectories are predicted. Subsequent works, such as GoRela \cite{gorela}, further improved this approach by sampling goal candidates along reachable lanes, filtering out unreachable areas, and achieving better alignment with the road geometry. Moreover, GoRela \cite{gorela} replaced NMS with a greedy selection strategy, resulting in more effective final goal set selection. %
However, this greedy sampling relies on a few hand-tuned hyperparameters.

To address the limitations of sparse goal candidate sampling, GoHome \cite{gohome} jointly predicts intermediate waypoints and the final goal point to guide the model toward feasible trajectories. However, this joint prediction approach often entangles errors between intermediate waypoints and goal estimation, leading to limited diversity and weaker performance in highly multi-modal scenarios. In contrast, DenseTNT \cite{densetnt} improves upon earlier goal-conditioned methods by densely sampling goal candidates over reachable areas and decoupling goal selection from trajectory prediction. By first identifying high-quality goal points and subsequently generating full trajectories conditioned on these goals, DenseTNT\cite{densetnt} achieves finer-grained coverage of the motion space, but this also increases computational effort. Furthermore, DenseTNT\cite{densetnt} only predicts one agent in the scene.

Regarding this shortcoming, we introduce our model HoliGraph:Goal that employs a learnable multi-stage goal selection, which guides future trajectories towards reachable parts of the road.
\section{Concept}
The architecture of the model is depicted in Fig.~\ref{fig:concept}.
It consists of data encoding, followed by an encoder-decoder GNN, which includes multi-stage goal selection and trajectory completion.
Our model predicts $K$ trajectories, hereafter referred to as modes, per agent.

\subsection{Data Representation}
The data used for prediction is organized in a heterogeneous graph, which is defined as $\mathcal{G = \{N, E}\}$, where $\mathcal{N}$ denotes the set of nodes and $\mathcal{E}$ denotes the set of edges with their corresponding edge features.
One sample consists of the dynamic context of the traffic participants, hereafter referred to as agents, and the road network given by the HD-Map.
The set of nodes $\mathcal{N = \{A,L,P,Q\}}$ consists of $4$ types:
\begin{itemize}
    \item Agent-nodes $\mathcal{A}$: A single agent-node $\pmb{a}_i^t$ refers to a measurement at time-step $t$ of the observed past trajectory of the \textit{i}-th agent and consists of the type and the current velocity in the agent-centric coordinate system: $\pmb{a}_i = (v_{x_i},v_{y_i},t_i)^\intercal$.
    \item Lane-nodes $\mathcal{L}$: A single lane-node $\pmb{l}_i$ refers to the origin (geometric center) of a lane $i$ in the hd-map, consisting of type and length, hence, $\pmb{l}_i = (t_i,len_i)^\intercal$.
    \item Point-nodes $\mathcal{P}$: A single point-node $\pmb{p}_i$ refers to a segment of a polyline belonging to a lane and holds information about type, length, and side (right, left, center), therefore, $\pmb{p}_i = (t_i,l_i,s_i)^\intercal$.
    \item Agent-query-nodes $\mathcal{Q}$: These are artificial nodes inserted at the same position as the agent-nodes of the last history timestep $T_h$.
    For every mode $k$ one agent-query-node is inserted for the respective agents. It is initialized with a learnable embedding of size $128$. %
\end{itemize}
\begin{figure}[h]
  \centering
  \includegraphics[width=0.8\columnwidth]{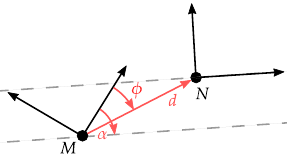}
  \caption{Schematic illustration of the relative edge features connecting 2 nodes $m$ and $n$.}
  \label{fig:relative_edges}
\end{figure}
The set of edges $\mathcal{E}$ contains information about the adjacency of different edge types $e$ as well as features belonging to those edges, which include spatial and temporal information; the latter is only available for agent-nodes.
Therefore, features $e_{mn}$ for an edge connecting to agent-nodes $a_i^{t_m}$ and $a_i^{t_n}$ of the same agent are calculated as follows:
\begin{equation}
    e_{mn} = \text{concat}\left[\sin\left(\alpha\right), \cos\left(\alpha\right),\sin\left(\phi\right),\cos\left(\phi\right), d,\Delta t_{kl}\right]
\end{equation}
with $\alpha$ being the angle between the two local coordinate systems $M$ and $N$, $\phi$ and $d$ describing the vector from $M$ to $N$ in frame $M$ in polar coordinates, depicted in Fig.~\ref{fig:relative_edges}.
$\Delta t_{mn}$ refers to the difference between the two time-steps and is only used when available.
The edge types $e$ will be introduced in the Sec.~\ref{sec:encoder}, \ref{sec:decoder}, and \ref{sec:goalsel}.
The naming scheme for edges connecting source-nodes to destination-nodes follows \textit{source-node, relation, destination-node}.

\subsection{Encoder}
\label{sec:encoder}
\begin{figure}[h]
  \centering
  \includegraphics[width=0.8\columnwidth]{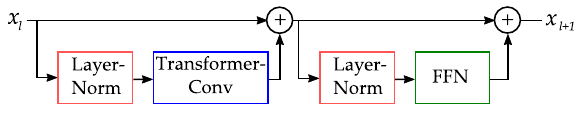}
  \caption{Architecture of a single Graph Attention layer.}
  \label{fig:gnn}
\end{figure}

In the encoder of the model, the initial node and edge features are first embedded in a hidden dimension of size $d_h = 128$.
\begin{equation}
    f_{\left\{\mathcal{N, E}\right\}} = \text{MLP} \left( \sum_i f_{\left\{\mathcal{N, E}\right\}}^i \right)
    \label{eq:embedding}
\end{equation}
Initial features of continuous values are embedded using a 2-layer MLP, resulting in node features $f_{\mathcal{N}}^c$ and edge features $f_{\mathcal{E}}^c$. 
Initial categorical features, i.e., type and side, are embedded using a simple lookup table with learnable embeddings in order to get $f_{\mathcal{N}}^t$ and $f_{\mathcal{N}}^s$.
Afterwards, embedded continuous and categorical features belonging to the same entity, i.e. edge or node, are summed and processed again by a 2-layer MLP, see Eq.~\ref{eq:embedding}.
Following the independent encoding of node features, we stack Graph Attention Layers, see Fig.~\ref{fig:gnn} to model interactions across different node types, allowing the model to capture heterogeneous agent dynamics.
A single Graph Attention Layer consists of a Transformer-Conv block \cite{graphtransformer}, followed by a Feed Forward Network (FFN) consisting of a 2-layer MLP using a hidden dimension of $512$.
The Transformer-Conv \cite{graphtransformer} takes the concept of Queries, Keys, and Values from a Transformer \cite{transformer} and applies it to graphs.
Instead of a fully-connected attention to all elements of a set, the attention is defined by the adjacency of an edge.
We use Layer-Norm to normalize the inputs to each block.
The encoder-GNN consists of a map-block and an agent-block.
The goal of the map-block is to generate meaningful lane features using the edge types:
\begin{itemize}
    \item \textit{(point, belongs-to, lane)}: Lane-nodes are updated to gather information about their boundaries and centerline, which are modeled as point-nodes.
    \item \textit{(lane, to, lane)}: Acknowledge lane-nodes about surrounding lane-nodes.
    For this purpose, a lane-node is connected to all lane-nodes within a radius of $125~\text{m}$.
    A categorical embedding describing the relation between the lane pair is added to the edge features.
    Potential relations are: \textit{successor, predecessor, right-neighbor, left-neighbor} and \textit{none}.
\end{itemize}
Distinguishing between lane-nodes and point-nodes has the benefit of a locally dense, but globally sparse representation of the map, reducing memory consumption and still allowing attention between distant parts of the road network. The map-block is executed only once and followed by the agent-block, which is executed twice and uses the following edge types:
\begin{itemize}
    \item \textit{(agent, suc, agent)}: Self-attention in the temporal dimension, i.e., agent-nodes yield information to successor agent-nodes.
    Typically, agent-nodes are connected with up to $20$ successor nodes.
    \item \textit{(agent, social, agent)}: Attention of agent-nodes to neighboring traffic participants.
    An agent-node is connected to other agent-nodes belonging to the same timestep within a radius of $50~\text{m}$.
    \item \textit{(lane, gives-traffic-info, agent)}: Update agent-nodes with map context.
    Hence, lane-nodes are connected to all agent-nodes in a radius of up to $50~\text{m}$.
\end{itemize}

\subsection{Decoder}
\label{sec:decoder}
Using the encoded heterogeneous graph provided by the encoder, the decoder's task is to generate meaningful agent-query features.
We use $4$ different edge types in a query-block to update the agent-query-nodes.
In total, two query-blocks are used.
\begin{itemize}
    \item \textit{(agent, self, agent-query)}: Attention to all agent-nodes belonging to the same agent.
    \item \textit{(agent, social, agent-query)}: Attention to neighboring agent-nodes at the last observed time-step $T_h$ closer than $100~\text{m}$.
    \item \textit{(lane, gives-traffic-info, agent-query)}: Attention to lane-nodes closer than $150~\text{m}$.
    \item \textit{(agent-query, self, agent-query)}: Attention to all other agent-query-nodes of the same agent in order to reduce mode collapse.
\end{itemize}
\subsection{Multi-stage goal selection}
\label{sec:goalsel}
The multi-stage goal selection is the main contribution of this work, as depicted in Fig.~\ref{fig:overview}.
The goal scoring is directly done in the heterogeneous graph, by predicting scores of edges connecting query-nodes with potential goals.
We distinguish between road-bound agents, i.e., agents that typically follow the road network like cars, trucks, motorcyclists, and cyclists, and non-road-bound agents, i.e., pedestrians. 
For road-bound agents, the task is to classify potential lane-nodes of $\mathcal{L}$ in the first stage.
The second stage enables a more fine-grained goal selection by classifying point nodes of $\mathcal{P_\text{rb}}$ within the most probable lane.
Therefore, the edges used for road-bound agents are as follows:
\begin{itemize}
    \item \textit{(agent-query, decide, lane)}: Road-bound agents are connected to all accessible lanes that they can reach while respecting the road network.
    \item \textit{(agent-query, decide, point)}: Having selected a goal lane, this edge connects the respective agent-query-node with all points belonging to that lane.
\end{itemize}
Since non-road-bound agents typically move more freely and travel shorter distances, a set of artificial \textit{point-nrb} nodes is generated around them with the task of directly selecting one of them as their future trajectory endpoint.
Therefore, we use the edge:
\begin{itemize}
    \item \textit{(agent-query, decide, point-nrb)}: Using the mean velocity of a non-road-bound agent, $8$ concentric circles around the query-node are generated. The radius of the $ith$ circle matches the distance travelled at mean velocity for $i$ seconds. To achieve a uniform point density, the number of points increases with the radius of the circle.
\end{itemize}
In order to estimate a probability distribution over all potential goals for a single agent-query node, we first calculate logits.
For one single logit $l_{i,j}$ connecting agent-query-node $i$ with goal $j$ we use the node features of the encoded heterogeneous graph, $f_i$, $f_j$ and edge feature $f_{i\rightarrow j}$:
\begin{equation}
    l_{i,j} = \text{MLP}\left(\text{concat}\left[f_i, f_j, f_{i\rightarrow j}\right]\right) + f_{i\rightarrow j}
\end{equation}
We group logits belonging to the same agent-query node $\pmb{q}_i$ and apply a softmax to yield the final scores $\pmb{s}_i$.
For every agent-query-node, we select the destination node belonging to the edge with the highest score as selected goal.
However, these goals do not fully cover the spatial $2$D space. Therefore, we employ a final goal regression afterwards.
We concatenate agent-query-node features, goal-node features, and respective edge features in order to calculate an offset, which is added to the position of the goal-node. %
\subsection{Trajectory Completion}
\label{sec:traj_complete}
The final trajectories are generated using a concatenation of agent-query-node features and the position of the regressed goal-points.
For each group (road-bound, non-road-bound), we use an MLP to predict a set of relative offsets $\delta \pmb{\mu}$ which yield the location $\pmb{\mu}$ by employing the cumulative sum of $\delta \pmb{\mu}$, along with a set of scales $\pmb{b}$ to form a Laplace distribution of the future trajectory.
For scoring of the trajectories, we use the score $\pmb{s}$ of the point goal classification.

\subsection{Loss}
\label{sec:loss}
We use a combination of $4$ losses.
For lane and point classification, we use a focal loss \cite{focalloss} $\mathcal{L_\text{lane}}$ and $\mathcal{L_\text{point}}$ each with $\alpha=0.75$ and $\gamma = 2$.
Goal regression uses a Huber loss $\mathcal{L}_\text{goal}$, and for trajectory completion, the negative log-likelihood $\mathcal{L}_\text{traj}$ is used.
It is beneficial to weight $\mathcal{L}_\text{traj}$ by a factor of $10$ more, resulting in the combined loss:
\begin{equation}
    \mathcal{L} = \mathcal{L}_\text{lane} + \mathcal{L}_\text{point} + \mathcal{L}_\text{goal} + 10 \, \mathcal{L}_\text{traj}
\end{equation}
Since $k$ modes are predicted, but only one ground-truth trajectory is available per agent, we impose a winner-takes-all strategy. A winner-mode is determined by stepwise selecting the mode performing best across the different tasks. Starting with lane classification, modes with their goal lane-node nearest to the ground-truth are winners. If more than one winner-mode is left, modes with their goal point-node nearest to the ground-truth are winners. If more than one winner-mode is left, modes with their goal-offset nearest to the ground-truth are winners. After that, a unique winner mode per agent is left, which is used for backpropagation.
\begin{figure*}[t]
    \centering
    \begin{subfigure}{0.4\textwidth}
        \centering
        \includegraphics[width=\columnwidth]{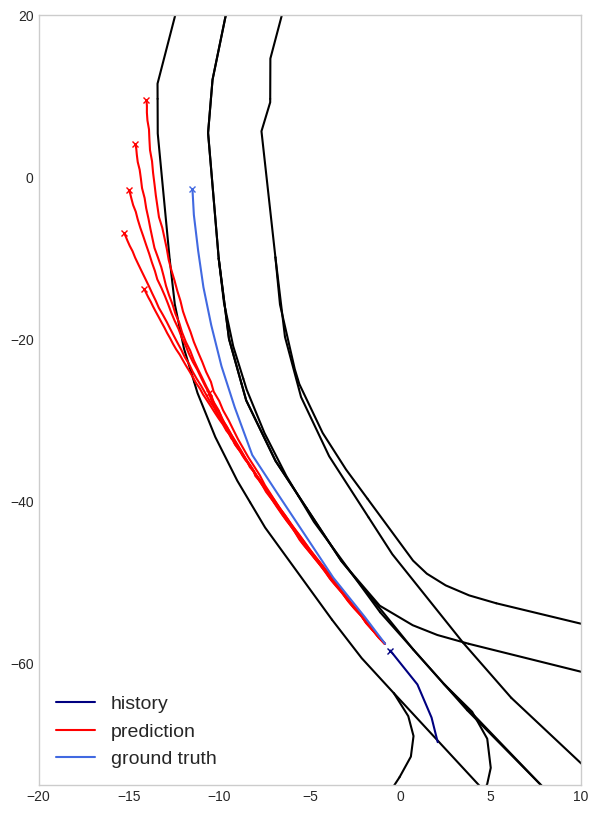}
        \caption{HoliGraph}
        \label{fig:holigraph}
    \end{subfigure}
    \hspace{20pt}
    \begin{subfigure}{0.4\textwidth}
        \centering
        \includegraphics[width=\columnwidth]{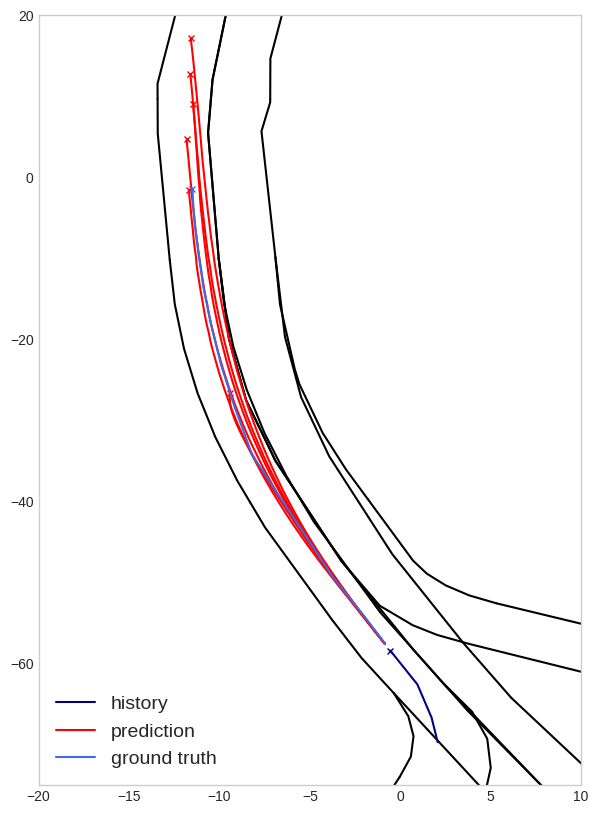}
        \caption{HoliGraph:Goal}
        \label{fig:holigraph_goal}
    \end{subfigure}
    \caption{Qualitative results for cross-evaluation. Models were trained on Argoverse2\cite{argoverse2} and predictions were made an NuScenes\cite{nuscenes} validation split. HoliGraph baseline fails to correctly attend to the road network resulting in Offroad-trajectories. HoliGraph:Goal however predicts trajectories more in compliance with the road-layout.}
    \label{fig:cross_eval}
\end{figure*}

\section{Evaluation}
\label{sec:eval}
In this section, we introduce the datasets Argoverse2 \cite{argoverse2} and NuScenes \cite{nuscenes} and the necessary adaptations to be used in cross-dataset evaluation, as well as the chosen Hyperparameters.
Afterwards, we present quantitative results proving the effectiveness of the multi-stage goal selection for improved generalization across datasets.
\subsection{Experimental settings}
\subsubsection{Datasets}
The Argoverse2 Motion Forecast Dataset \cite{argoverse2} contains $763~\text{h}$ of data resulting in a large collection of $199908/24988$ train/val samples, each with a duration of $11~\text{s}$, split up in $5~\text{s}$ history and $6~\text{s}$ future.
Regarding the evaluation, up to $K=6$ trajectories per agent are allowed.
The data was collected at $10~\text{Hz}$ in six U.S. cities with complex, unique driving environments: Miami, Austin, Washington DC, Pittsburgh, Palo Alto, and Detroit.
The NuScenes Dataset \cite{nuscenes} contains $92~\text{h}$ of data with $32186/9041$ train/val samples.
It was collected in Boston and Singapore at $2~\text{Hz}$.
Each sample contains $2~\text{s}$ of history and $6~\text{s}$ of future.
Regarding the evaluation, up to $K=10$ trajectories per agent are allowed.
To be able to use both datasets with the same model, we made adaptations to the NuScenes\cite{nuscenes} data.
Our Model runs at $10~\text{Hz}$, meaning input and output trajectories are treated as $10~\text{Hz}$ data. 
In order to use NuScenes\cite{nuscenes} for cross-generalization, Unitraj \cite{unitraj} upscaled the dataset to $10~\text{Hz}$.
We follow this schema and use a linear interpolation of 2 consecutive waypoints to fill the resulting gaps of $0.4~\text{s}$.

\subsubsection{Hyperparameters}
We train our model using AdamW Optimizer \cite{adamw} with initial $\beta_1 = 0.9$ and $\beta_2 = 0.95$.
The learning rate is gradually increased during a warmup epoch from $0$ up to $5e-4$, and a cosine scheduler is used, decreasing the learning rate back to $0$ over $39$ epochs.
We use a weight decay of $1e-4$ and a total batch size of $64$.
LeakyReLU is used as an activation.
The model is trained using $2$ RTX6000 Ada GPUs, resulting in a training time of approx. $30~\text{h}$ for Argoverse2 and $8.6\text{h}$ for Nuscenes \cite{nuscenes}.
During training, we use dropout of $0.1$ and employ data augmentation, which scales the scene by a random factor $\sigma \in [0.8; 1.2]$ and removes $10$\% of the agents in the scene.
We use $128$ as hidden dimensions for all node and edge features.
A forward pass for a single traffic scene takes on average $69~\text{ms}$, resulting in real-time capabilities.
\subsubsection{Metrics}
We use a wide variety of metrics.
Depending on the task, we report Minimum Average Displacement Error ($\text{minADE}_K$), Minimum Final Displacement Error ($\text{minFDE}_K$), Minimum Missrate ($\text{minMR}_K$).
The displacement is calculated using the $l^2$ norm between predicted and ground-truth positions. 
$\text{minMR}_K$ measures the proportion of trajectories with $\text{$\text{minFDE}_K$} > 2~\text{m}$.
NuScenes calculates the miss rate differently, calling it $\text{MissRateTopK\_2}_K$.
A trajectory is considered a miss, if any waypoint has a displacement error $> 2~\text{m}$.
Furthermore, Brier Minimum Final Displacement Error ($\text{b-minFDE}_K$) is reported, using the score of the selected trajectory to add $\left(1 - s^2\right)$ to the base metric.
The Offroad Rate (ORR) measures the proportion of trajectories that have at least one waypoint of their future trajectories outside of the lanes given by the HD-map. Metrics with the subscript $K$ are only evaluated on the $K$ highest-scoring trajectories.
\begin{table}[h]
    \centering
    \caption{Results on Focal Agent Argoverse2\cite{argoverse2} validation split in comparison to SotA models where the results are from Unitraj\cite{unitraj}.}
    \begin{tabular}{@{}l  c c c c @{}}
         \toprule
         Model & $\text{b-minFDE}_{6}$ & $\text{minFDE}_{6}$ & $\text{minADE}_{6}$ & $\text{minMR}_{6}$ \\ \midrule
         Wayformer\cite{wayformer} &  2.38~m & 1.75~m & 0.85~m & 0.28 \\
         Autobot\cite{autobot} &  2.51~m & 1.70~m & 0.85~m & 0.27 \\
         MTR\cite{mtr} &  \textbf{2.08~m} & 1.68~m & 0.85~m & 0.30 \\ \midrule
         HoliGraph &  \underline{2.15~m} & \textbf{1.52~m} & \textbf{0.74~m} & \underline{0.21} \\
         HoliGraph:Goal &  2.24~m & \underline{1.55~m} & \underline{0.76~m} & \textbf{0.20} \\
         \bottomrule
    \end{tabular}
    \label{tab:argo2}
\end{table}
\begin{table*}[t]
    \centering
    \caption{Results on NuScenes\cite{nuscenes} benchmark in comparison to SotA models.}
    \begin{tabular}{@{}l  c c c c c c@{}}
         \toprule
         Model & $\text{minADE}_{5}\downarrow$ & $\text{minADE}_{10}\downarrow$ & $\text{MissRateTopK\_2}_5\downarrow$ & $\text{MissRateTopK\_2}_{10}\downarrow$ & $\text{minFDE}_{1}\downarrow$ & ORR$\downarrow$ \\ \midrule
         PGP \cite{deo2022multimodal} & \textbf{1.30~m} & \textbf{1.00~m} & 0.61 & \textbf{0.37} & \underline{7.17~m} & \underline{0.03} \\
         LaPred \cite{kim2021lapred} & 1.47~m & 1.12~m & \textbf{0.53} & 0.46 & 8.37~m & 0.09 \\
         Thomas \cite{gilles2021thomas} & \underline{1.33~m} & \underline{1.04~m} & \underline{0.55} & \underline{0.42} & \textbf{6.71~m} & \underline{0.03} \\
         Autobot \cite{autobot} & 1.37~m & 1.07~m & 0.62 & 0.44 & 8.19~m & \underline{0.03} \\ \midrule
         HoliGraph & 1.73~m & 1.14~m & 0.58 & 0.47 & 8.71~m & 0.07 \\
         HoliGraph:Goal & 1.72~m & \underline{1.04~m} & 0.65 & \underline{0.42} & 10.37~m & \textbf{0.02} \\
         \bottomrule
    \end{tabular}
    \label{tab:nusc}
\end{table*}
\begin{table*}[t]
    \centering
    \caption{Results for the cross-dataset generalization. The first row contains the results of training and evaluation on NuScenes\cite{nuscenes}. The second row contains results for training on Argoverse2\cite{argoverse2}(Av2) and evaluation on NuScenes \cite{nuscenes}. The third row measures the relative drop in performance when moving to the cross-dataset setting. Results for SotA models are taken from Unitraj\cite{unitraj}.}
    \begin{tabular}{@{}l c c c c c c c c c c@{}}
         \toprule
         Model & Train & Eval & $\text{b-minFDE}_{6}\downarrow$ & $\text{minFDE}_{6}\downarrow$ & $\text{minADE}_6\downarrow$ & $\text{minMR}_{6}\downarrow$ & $\text{ORR}\downarrow$ \\ \midrule
         Wayformer \cite{wayformer} & Nusc & Nusc & 3.06~m & 2.50~m & 1.04~m & 0.42 & - \\
          & Av2 & Nusc & 3.69~m & 3.07~m & 1.44~m & 0.48 & - \\
         \cline{4-8}
         \rule{0pt}{2.0ex} & & & 20.6\% & 22.8\% & 38.5\% & 14.3\% & - \\\midrule
         Autobot \cite{autobot} & Nusc & Nusc & 3.36~m & 2.62~m & 1.21~m & 0.40 & -\\
          & Av2 & Nusc & 4.35~m & 3.52~m & 1.59~m & 0.49 & -\\
          \cline{4-8}
          \rule{0pt}{2.0ex}& &  & 29.5\% & 34.4\% & 31.4\% & 22.5\% & - \\\midrule
         MTR \cite{mtr} & Nusc & Nusc & 2.86~m & 2.33~m & 1.06~m & 0.41 &  - \\
          & Av2 & Nusc & 3.72~m & 3.10~m & \underline{1.42~m} & 0.47 & - \\
          \cline{4-8}
          \rule{0pt}{2.0ex}& &  & 30.1\% & 33\% & 34\% & 14.6\% & - \\\midrule
         HoliGraph & Nusc & Nusc & 3.09~m & 2.49~m & 1.27~m & 0.38 & 0.08 \\
          & Av2 & Nusc & \underline{3.42~m} & \underline{2.87~m} & 1.46~m & \underline{0.41} & \underline{0.10} \\
          \cline{4-8}
          \rule{0pt}{2.0ex}&  &  & 10.7\% & 15.3\% & 15\% & 7.9\% & 25\% \\\midrule
         HoliGraph:Goal & Nusc & Nusc & 3.13~m & 2.43~m & 1.25~m & 0.37 & 0.03 \\
          & Av2 & Nusc & \textbf{3.38~m} & \textbf{2.68~m} & \textbf{1.32~m} &  \textbf{0.39} & \textbf{0.03} \\
          \cline{4-8}
         \rule{0pt}{2.0ex}&  &  & 8\% & 10.3\% & 5.6\% & 5.4\% & 0\% \\
         \bottomrule
    \end{tabular}
    \label{tab:nusc_cross}
\end{table*}
\subsection{Results}
We evaluated two versions of our model: HoliGraph:Goal, referring to the presented model, and HoliGraph, where the multi-stage goal selection is removed and instead trajectories and their corresponding scores are directly predicted using agent-query-node features.
Tab.~\ref{tab:argo2} shows the results on the Argoverse2\cite{argoverse2} validation split using their official metrics. 
The two models perform similarly; however, HoliGraph:Goal shows a drop in performance, regarding $\text{b-minFDE}_6$. 
Since $\text{b-minFDE}_6$ is approximately equal, this indicates that the performance drop originates from the predicted scores of the trajectories and not the trajectories themselves.
However, the results are still competitive and on par with SotA models.
Tab.~\ref{tab:nusc} depicts the results on the NuScenes\cite{nuscenes} validation split, using official NuScenes metrics.
Here HoliGraph:Goal shows a clear increase in performance regarding all $10$ predicted trajectories per agent ($\text{minADE}_{10}$, $\text{MissRateTopK\_2}_{10}$, ORR), especially a decrease in ORR from $7\%$ to $2\%$.
With regard to the size of the two datasets, i.e., Argoverse2\cite{argoverse2} being 6 times the size of NuScenes\cite{nuscenes}, it can be concluded that multi-stage goal selection is advantageous when less training data is available.
Metrics that include the scoring of trajectories, i.e. $\text{minADE}_{5}$, $\text{MissRateTopK\_2}_{5}$, $\text{minFDE}_{1}$, show the same behavior as on Argoverse2 \cite{argoverse2}. 
This means that the performance decreases slightly, which indicates that the scoring of trajectories could also be improved.
Despite this, HoliGraph:Goal outperforms other SotA models regarding ORR.
\subsection{Cross-dataset evaluation}
For cross-dataset evaluation, we follow the evaluation scheme of Unitraj \cite{unitraj} in order to compare to their results.
This means, that we report $\text{b-minFDE}_6$, $\text{minFDE}_6$, $\text{minADE}_6$, $\text{minMR}_6$ and extend their metrics by ORR.
As demonstrated in SceneAttack \cite{scene_transformer}, ORR is an important metric to measure generalization, since it not only takes into account the best trajectory per agent, but all of them.
Therefore, we also report it.
In Tab.~\ref{tab:nusc_cross}, the results on the NuScenes \cite{nuscenes} validation split are reported twice for every model.
Firstly, when trained on NuScenes\cite{nuscenes} and afterwards when trained on Argoverse2\cite{argoverse2}, also including the relative change between the two datasets.
Results for other models are taken from Unitraj \cite{unitraj}.
HoliGraph already reduces the relative change between the two datasets, which can be assumed to be due to translational and rotational invariant modeling of the traffic scene.
HoliGraph:Goal further reduces the performance drop between the datasets, yielding the best overall results for cross-dataset evaluation.
This indicates that the proposed multi-stage goal selection enhances the model's road awareness and helps avoid off-road predictions, leading to better generalization, as can be seen in Fig.~\ref{fig:cross_eval}.
\section{Conclusion}
In this work, we proposed a new multi-stage goal selection process for Trajectory Prediction.
The method is based on a heterogeneous Graph and distinguishes between road-bound and non-road-bound agents to accommodate their specific motion characteristics.
Trajectories are guided towards those goals, leading to lower off-road rates and an overall better generalization to unseen data, as shown in the cross-dataset evaluation Tab.~\ref {tab:nusc_cross}.
On smaller datasets, the multi-stage goal selections increase the accuracy of the predicted waypoints of the trajectory, as shown in Tab.~\ref{tab:nusc}. %
However, it should be mentioned that the scoring of the predicted trajectories could be further optimized.
From the results, it can be assumed that instead of simply using the score of the point goal classification, a separate scoring module, just as in the HoliGraph baseline, could increase scoring performance and should therefore be tested in future work.
Additionally, the selected goals could be used to introduce a drivability loss, comparing the heading of the trajectory with the heading of the selected lanes, to not only train on the mode closest to the ground-truth, but to train on all predicted modes, leading to even better generalization.
\section*{Acknowledgment}
This paper was created in the "Country 2 City - Bridge" project of the "German Center for Future Mobility", which is funded by the German Federal Ministry of Transport.
Responsibility for the information and views set out in this publication lies entirely with the authors.

\bibliographystyle{IEEEtran}
\bibliography{library}

\end{document}